\documentclass{article}
\usepackage{spconf,amsmath,graphicx,hyperref}
\usepackage{amssymb,bm,amsthm}
\usepackage{float}
\usepackage{booktabs}
\usepackage{tabularx}
\usepackage{enumitem}
\usepackage{bm}
\usepackage{xcolor}

\usepackage{flushend} 

\usepackage{math_symbols}

\title{Bayesian Jammer Localization with a Hybrid CNN and Path-Loss Mixture of Experts}

\name{
Mariona Jaramillo-Civill\textsuperscript{1},
Luis González-Gudiño\textsuperscript{2},
Tales Imbiriba\textsuperscript{2},
Pau Closas\textsuperscript{1}\thanks{This work has been partially supported by the National Science Foundation under Award 1845833 and 2326559 and 2530870.}}

\address{
\textsuperscript{1}Dept. of Electrical \& Computer Engineering, Northeastern University, Boston, MA, USA \\
\textsuperscript{2}Dept. of Computer Science, University of Massachusetts Boston, Boston, MA, USA}

\begin{document}
\ninept
\maketitle

\begin{abstract}
Global Navigation Satellite System (GNSS) signals are vulnerable to jamming, particularly in urban areas where multipath and shadowing distort received power. Previous data-driven approaches achieved reasonable localization but poorly reconstructed the received signal strength (RSS) field due to limited spatial context.  
We propose a hybrid Bayesian mixture-of-experts framework that fuses a physical path-loss (PL) model and a convolutional neural network (CNN) through log-linear pooling. The PL expert ensures physical consistency, while the CNN leverages building-height maps to capture urban propagation effects. Bayesian inference with Laplace approximation provides posterior uncertainty over both the jammer position and RSS field. Experiments on urban ray-tracing data show that localization accuracy improves and uncertainty decreases with more training points, while uncertainty concentrates near the jammer and along urban canyons where propagation is most sensitive.
\end{abstract}


\section{Introduction}
\label{sec:intro}

Global Navigation Satellite System (GNSS) signals have become indispensable for civilian and critical infrastructure applications, including autonomous driving, aviation, telecommunications, and precise timing~\cite{morton_position_2021,amin_vulnerabilities_2016}. This reliance increases vulnerability to interference: low-cost L-band jammers can overpower receivers and disrupt services, while unintentional sources such as malfunctioning electronics or DME systems further increase interference risks~\cite{borio_gnss_2012,li_dual-domain_2019}.

Given these threats, detecting, localizing, and mitigating interference sources is essential. Traditional monitoring relies on costly equipment and human supervision~\cite{mehr_navigation_2023}. A scalable alternative leverages crowdsourced data: mobile users report received signal strength (RSS) across an area, allowing a central or distributed system to infer the propagation field and estimate jammer positions~\cite{borio_jammer_2016,arias-de-reyna_crowd-based_2018,jaramillo-civill_jammer_2025}. With sufficient participation, such collaborative inference enables accurate and timely jammer localization.

Prior work has explored several approaches~\cite{borio_jammer_2016,olsson_using_2023,bai_map_2024,bhamidipati_simultaneous_2018}. Path-loss (PL) models offer physical interpretability but fail in dense urban areas dominated by multipath and shadowing, whereas purely data-driven neural networks (NNs) can fit complex patterns but extrapolate poorly and lack physical grounding. Augmented Physics-Based Models (APBMs)~\cite{jaramillo-civill_jammer_2025,nardin_jamming_2023,nardin_crowdsourced_2023} combine both by correcting the PL equation with a NN to model non-ideal propagation. Although effective, APBMs still struggle in urban settings where blockage, multipath, and sparse observations hinder accurate field reconstruction. Recent deep learning approaches~\cite{herzalla_graph_2025} introduce relational structure using graph attention networks with centroid priors~\cite{wang_efficient_2020}, capturing spatial dependencies but remaining purely data-driven and lacking physical interpretability \cite{closas_emerging_2024}.
Our previous work~\cite{jaramillo-civill_jammer_2025} localized jammers by fitting RSS fields with a NN to initialize the APBM, which improved position estimates but yielded poor field reconstruction due to the NN’s limited spatial awareness and lack of probabilistic grounding. 

To address these limitations, we propose a hybrid Bayesian mixture-of-experts framework that fuses a path-loss expert with a convolutional neural network (CNN) through log-linear pooling. The CNN leverages building-height maps to capture multipath and shadowing effects, while the path-loss expert enforces physical consistency. A data-dependent prior over the jammer position, inspired by weighted centroid methods~\cite{wang_efficient_2020}, provides theoretically grounded localization and principled uncertainty quantification.  
Prior probabilistic approaches have addressed uncertainty, such as Gaussian mixture probability hypothesis density filters for multi-jammer tracking~\cite{bhamidipati_simultaneous_2018} and Gaussian process regression for power-map reconstruction~\cite{bai_map_2024}. While effective, these methods yield only indirect uncertainty over RSS values and lack explicit probabilistic modeling of the jammer position and propagation field. In contrast, our formulation provides explicit posterior uncertainty over both the jammer position and propagation field, a key capability for future active learning strategies.

The remainder of the article is organized as follows. Section~\ref{sec:problem} introduces the problem formulation, Section~\ref{sec:methodology} describes the hybrid Bayesian mixture-of-experts model and the corresponding inference.  Section~\ref{sec:results} presents the simulation framework and reports experimental analysis, followed by concluding remarks in Section~\ref{sec:conclusion}.

\section{Problem Formulation}
\label{sec:problem}

We aim to jointly estimate the interference propagation field and jammer position from noisy received signal strength (RSS) measurements. Let $\mathcal{G}\subset\mathbb{Z}^2$ denote a point in a discrete grid of size $H\times W$. We observe:
$\mathcal{D}=\{(\mathbf{p}_i,y_i)\}_{i=1}^N$, where $\mathbf{p}_i\in\mathcal{G}$ specifies the grid location of receiver $i$ and $y_i\in\mathbb{R}$ (dBW) is the corresponding RSS measurement~\cite{dardari_indoor_2015}. In addition, we assume access to a building-height map $h:\mathcal{G}\to\mathbb{R}$ covering the entire grid, which provides side information about the propagation context at every location. We denote by $\mathbf{H}_{\text{bld}}=h(\mathcal{G})\in\mathbb{R}^{H\times W}$ the corresponding height raster used as model input.

The measurements are generated by an unknown propagation function 
$f_{\mathrm{TRUE}}(\mathbf{p},\mathbf{H}_{\text{bld}};\bftheta)$ 
that depends on the jammer position $\bftheta$ and the environment $\mathbf{H}_{\text{bld}}$, 
subject to additive noise $\xi_i$:
\begin{equation}
y_i = f_{\mathrm{TRUE}}(\mathbf{p}_i,\mathbf{H}_{\text{bld}};\bftheta) + \xi_i,\quad i=1,\dots,N.
\end{equation}

In its most simple case, the measurement model follows the well-known path-loss propagation model. However, due to unmodeled phenomena such as multipath, shadowing, or signal blockage, this model is generally inaccurate. This makes the actual $f_{\mathrm{TRUE}}$ to be generally unknown.
Therefore, we model the observed RSS as
\begin{equation}\label{eq:muRSS}
y_i = \mu(\bfp_i;\mathbf{H}_{\text{bld}},\bfpsi) + \xi_i,\quad \xi_i\sim\mathcal N(0,\beta^{-1}),\;\; i=1,\dots,N,
\end{equation}
where $\mu:\;\mathcal{G}\times\mathbb{R}^{H\times W}\times\bfPsi \;\to\; \mathbb{R}$
is a parametric mean function that depends on the receiver position $\bfp_i$, 
the building-height raster $\mathbf{H}_{\text{bld}}\in\mathbb{R}^{H\times W}$, 
and parameters $\bfpsi\in\bfPsi$ including the jammer position $\bftheta\in\mathbb R^2$. 
Thus, $\mu$ defines the model of the RSS field, while the Gaussian noise term $\xi_i$ with inverse variance $\beta$ accounts for additive measurement noise.
Bayesian inference places a prior on $\bfpsi$ and provides the posterior $p(\bfpsi\mid\mathcal D)$ over parameters.

The approximation to $f_{\mathrm{TRUE}}$ is then given by the posterior predictive distribution
\begin{equation}\label{eq:pred_gen}
p(y^\star\mid \mathbf p^\star,\mathcal D) \;=\; 
\int \mathcal N\!\big(y^\star \mid \mu (\mathbf p^\star;\mathbf{H}_{\text{bld}},\bfpsi),\beta^{-1}\big)\,
p(\bfpsi\mid\mathcal D)\,d\bfpsi.
\end{equation}

The learning task is therefore twofold:  
(i) \emph{field prediction:} obtain the posterior predictive distribution to estimate the RSS at unobserved positions $\mathbf p^\star\in\mathcal G\setminus\{\bfp_i\}$; and  
(ii) \emph{jammer localization:} extract the marginal posterior $p(\bftheta\mid\mathcal D)$, which provides both an estimate of the jammer location and its uncertainty.

\section{Joint estimation of jammer's location and signal propagation model}
\label{sec:methodology}

This section first presents the proposed hybrid physics-based and data-driven modeling of the RSS over space, that is function $\mu(\cdot)$ in \eqref{eq:muRSS}. Then, a Laplace approximation is taken in order to obtain closed form estimators of both the jammer's location and the model parameters. Finally, the section provides means to reconstruct the RSS field over unobserved points in the area of interest.

\subsection{Expert Models, Log-linear Pooling and Priors}

We consider two \textit{competing} models, one based on physics and another on data:

\noindent\textbf{\emph{1. Path-loss expert.}} 
The widely used log-distance model~\cite{goldsmith_wireless_2005} is
\begin{equation}
\mu^{\mathrm{PL}}(\mathbf p; \bftheta,P_0,\gamma)=
P_0 - 10\,\gamma\,\log_{10}\!\big(\|\mathbf p-\bftheta\|_2+\varepsilon\big),
\end{equation}
where $\bftheta\in\mathbb R^2$ is the jammer position, $P_0$ the transmit power at 1\,m (dBW), 
$\gamma$ the path-loss exponent, and $\varepsilon>0$ avoids singularities. While this is accurate in open-sky environments, shadowing and reflections render it unreliable in more complex situations.

\begin{figure}[t]
    \centering
    \includegraphics[width=0.85\linewidth]{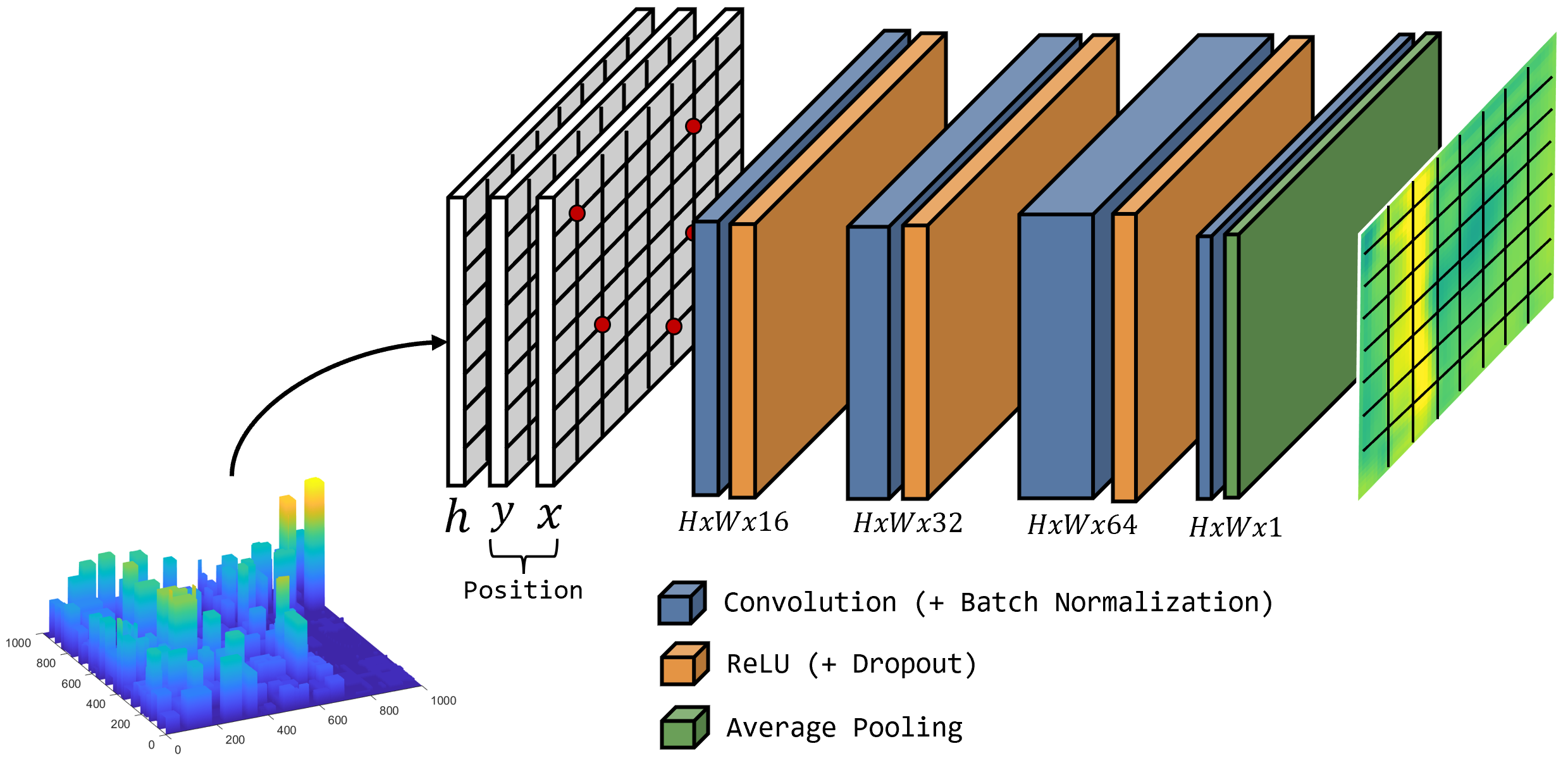}
    \vspace{-0.5cm}
    \caption{CNN expert architecture. The input $\mathbf{X}\in\mathbb R^{3\times H\times W}$ 
    combines building heights and spatial coordinates. 
    The network consists of convolutional layers with batch normalization and ReLU, with dropout regularization, followed by a final convolution and global average pooling to produce an $H\times W$ RSS field prediction.}
    \label{fig:cnn_architecture}
\end{figure}

\noindent\textbf{\emph{2. CNN expert.}} We consider a convolutional network (Fig.~\ref{fig:cnn_architecture}), with parameters $\bfomega$ acting on the grid.
\begin{equation}
g_{\bfomega}:\;\mathbb R^{3\times H\times W}\to \mathbb R^{H\times W}, 
\quad 
\mu^{\mathrm{CNN}}(\mathbf p; \bfX,\bfomega) = \delta_{\mathbf p}(g_{\bfomega}(\bfX)).
\end{equation}
Let $\bfX=[\mathbf{H}_{\text{bld}},\mathbf{P}_x,\mathbf{P}_y]\in\mathbb R^{3\times H\times W}$ be the three–channel input grid, where $\mathbf{H}_{\text{bld}}$ encodes building heights and $(\mathbf{P}_x,\mathbf{P}_y)$ encode spatial coordinates.
For a matrix $\bfM\in\mathbb R^{H\times W}$, we define the indexing operator
\begin{equation}
\delta_{\mathbf p}:\mathbb R^{H\times W}\to\mathbb R,
\quad 
\delta_{\mathbf p}(\bfM)=\bfM[\mathbf p],
\end{equation}
which extracts the entry of $\bfM$ at grid location $\mathbf p\in\mathcal G$.

\noindent\textbf{\emph{Log-linear pooling.} } 
For each expert, the observed RSS is modeled as Gaussian:
\begin{align}
y_i|\bfomega &\sim \mathcal N(\mu^{\mathrm{CNN}}(\bfp_i; \bfX,\bfomega),\,\beta_1^{-1}),\\
y_i | \bftheta,P_0,\gamma &\sim \mathcal N(\mu^{\mathrm{PL}}(\bfp_i;\bftheta,P_0,\gamma),\,\beta_2^{-1}),
\end{align}
with precisions $\beta_1,\beta_2>0$ respectively.

We combine those Gaussian experts via log-linear pooling~\cite{koliander_fusion_2022} 
with a mixing weight $\lambda\!\in[0,1]$ controlling their relative influence.
Unlike the parameters $\boldsymbol{\psi}=(\boldsymbol{\omega},\bftheta,P_0,\gamma)$, which are modeled as random following a Bayesian approach, $\lambda$ is estimated deterministically under a type-II maximum likelihood scheme.
The resulting pooled Gaussian has mean:
%
\begin{equation}
\begin{aligned}
\mu(\bfp_i;\mathbf{H}_{\text{bld}},\bfpsi)
&= \beta^{-1}\!\Big(
   \lambda\beta_1\,\mu^{\mathrm{CNN}}(\bfp_i;\bfX,\bfomega) \\
&\qquad\quad + (1-\lambda)\beta_2\,\mu^{\mathrm{PL}}(\bfp_i;\bftheta,P_0,\gamma)
\Big).
\end{aligned}
\end{equation}
and precision $\beta=\lambda\beta_1+(1-\lambda)\beta_2$. Therefore, assuming i.i.d.\ observations, the pooled likelihood is:
\begin{equation}
p(\mathcal{D}|\bfpsi)=
\prod_{i=1}^N \mathcal N\!\big(y_i; \mu(\bfp_i;\mathbf{H}_{\text{bld}},\bfpsi),\,\beta^{-1}\big) \;.
\end{equation}

\noindent\textbf{\emph{Priors.}}
We assume independent priors over the full parameter vector $\bfpsi$; used to encode physical knowledge and regularize the CNN:
\begin{align}
p(\bfomega) &= \mathcal{N}(\bfomega; \mathbf{0},\,\sigma_{\bfomega}^2 \mathbf{I}), \\
p(P_0) &= \mathcal{N}\!\left(P_0\;;\;\tfrac{P_{\min}+P_{\max}}{2},\,\Big(\tfrac{P_{\max}-P_{\min}}{2}\Big)^2\right), \label{eq:prior_p0}\\
p(\gamma) &= \mathcal{N}\!\left(\gamma\;;\,\;\tfrac{\gamma_{\min}+\gamma_{\max}}{2},\,\Big(\tfrac{\gamma_{\max}-\gamma_{\min}}{2}\Big)^2\right), \label{eq:prior_gamma}\\
p(\bftheta) &= \mathcal{N}(\bftheta;\, \bfmu_c,\,\sigma_c^2\mathbf{I}), \quad 
\bfmu_c = \frac{\sum_{i=1}^N \exp(y_i/\tau)\,\bfp_i}{\sum_{i=1}^N \exp(y_i/\tau)}. \label{eq:prior_omega}
\end{align}
Here $P_0\in[P_{\min},P_{\max}]$, $\gamma\in[\gamma_{\min},\gamma_{\max}]$, 
and the prior on $\bftheta\in\mathbb R^2$ is centered at a temperature-weighted centroid of the measurements, 
controlled by $\tau$, constituting a data-dependent prior~\cite{darnieder_bayesian_2011}.
Assuming independence, the joint prior factorizes as
\begin{equation}
p(\bfpsi)=p(\bfomega)\,p(P_0)\,p(\gamma)\,p(\bftheta)\;.
\end{equation}

\subsection{Posterior Inference and Laplace Approximation}
Applying Bayes’ rule, the posterior over the parameters is
\begin{equation}
p(\bfpsi\mid\mathcal{D}) \;\propto\; p(\mathcal{D}|\bfpsi)\,p(\bfpsi),
\end{equation}
with MAP estimate $\hat{\bfpsi}_{\mathrm{MAP}}=\arg\max p(\bfpsi\mid\mathcal{D})$ which can be equivalently obtained by minimizing the negative log-posterior: 
\begin{equation}
\mathcal{J}(\bfpsi) \;\propto\;
\frac{\beta}{2}\sum_{i=1}^N\!\big(y_i-\mu(\bfp_i;\mathbf{H}_{\text{bld}},\bfpsi)\big)^2
+\mathcal{R}(\bfpsi),
\label{eq:log_posterior}
\end{equation}
where $\mathcal{R}(\bfpsi)=-\log p(\bfomega)-\log p(P_0)-\log p(\gamma)-\log p(\bftheta)$.  

To make posterior inference tractable, we approximate the posterior around $\hat{\bfpsi}_{\mathrm{MAP}}$ with a Gaussian via Laplace’s method~\cite{bishop_pattern_2006}:
\begin{equation}\label{eq:LaplaceApprox}
p(\bfpsi\mid\mathcal{D}) 
\;\approx\; \mathcal{N}\!\big(\bfpsi\;\big|\;\hat{\bfpsi}_{\mathrm{MAP}},\,\mathbf{H}^{-1}\big), 
\quad 
\mathbf{H}=\nabla^2_{\bfpsi}\mathcal{J}(\bfpsi)\big|_{\bfpsi=\hat{\bfpsi}_{\mathrm{MAP}}}.
\end{equation}
Computing the exact Hessian is costly for large CNNs, so we adopt the Gauss--Newton approximation~\cite{mackay_information_2019}:
\begin{align}
\mathbf{H} 
&\;\approx\; \beta\,\mathbf{J}^\top \mathbf{J} \;+\; \nabla_{\bfpsi}^2 \mathcal{R}(\hat{\bfpsi}_{\mathrm{MAP}}), \\[6pt]
\mathbf{J} 
&= \left( \dots, \nabla_{\bfpsi}\,\mu(\bfp_i;\mathbf{H}_{\text{bld}},\hat{\bfpsi}_{\mathrm{MAP}}), \dots \right)^\top, \\[6pt]
\nabla_{\bfpsi}^2 \mathcal{R}(\hat{\bfpsi}_{\mathrm{MAP}}) 
&= \mathrm{blockdiag}\!\Big(
\sigma_{\bfomega}^{-2} \mathbf{I},\;
\sigma_{P_0}^{-2},\;
\sigma_\gamma^{-2},\;
\sigma_c^2 \mathbf{I}
\Big).
\end{align}
Partitioning $\bfpsi=(\bftheta,\bfvarphi)$, with $\bfvarphi=(\bfomega,P_0,\gamma)$, the Laplace covariance $\mathbf{H}^{-1}$ can be block-partitioned using the Schur complement, and the marginal posterior of the jammer's location $\bftheta$ is
\begin{equation}
p(\bftheta\mid\mathcal{D})\approx
\mathcal{N}\!\big(\bftheta\mid \hat{\bfpsi}_{\mathrm{MAP}_{\bftheta}},\,\bfSigma_{\bftheta}\big), 
\bfSigma_{\bftheta}=\big(\bfH_{\bftheta\bftheta}-\bfH_{\bftheta\varphi}\bfH_{\varphi\varphi}^{-1}\bfH_{\varphi\bftheta}\big)^{-1} \;,
\end{equation}
where $\hat{\bfpsi}_{\mathrm{MAP}_{\bftheta}}$ denotes the $\bftheta$ component of  $\hat{\bfpsi}_{\mathrm{MAP}}$.

\subsection{Predictive Distribution}
Similar to \eqref{eq:pred_gen}, at an unobserved location $\mathbf{p}'$, the predictive distribution of the RSS is
\begin{equation}
p(y'\mid\mathbf p',\mathcal{D})=
\int \mathcal N\!\big(y'; \mu(\mathbf p';\mathbf{H}_{\text{bld}},\bfpsi),\beta^{-1}\big)\,
p(\bfpsi\mid\mathcal D)\,d\bfpsi \;,
\end{equation}
which can manipulated using the Laplace in \eqref{eq:LaplaceApprox} and linearizing
$\mu(\mathbf p';\mathbf{H}_{\text{bld}},\bfpsi)$ around $\hat{\bfpsi}_{\mathrm{MAP}}$,
\begin{align}
\mu(\mathbf p';\mathbf{H}_{\text{bld}},\bfpsi)
&\approx 
\mu(\mathbf p';\mathbf{H}_{\text{bld}},\hat{\bfpsi}_{\mathrm{MAP}})
+ \mathbf g^\top(\bfpsi-\hat{\bfpsi}_{\mathrm{MAP}}), \nonumber\\
\mathbf g
&= \nabla_{\bfpsi}\,\mu(\mathbf p';\mathbf{H}_{\text{bld}},\hat{\bfpsi}_{\mathrm{MAP}})
\end{align}
resulting in the Gaussian approximation of the distribution
\begin{equation}
\label{eq:pred_dist}
p(y'\mid\mathbf p',\mathcal D)\;\approx\;
\mathcal N\!\Big(y';\;\mu(\mathbf p';\mathbf{H}_{\text{bld}},\bfpsi_\mathrm{MAP}),\;
\beta^{-1}+\mathbf g^\top \mathbf H^{-1}\mathbf g\Big).
\end{equation}
The predictive variance decomposes into an aleatoric term 
$\beta^{-1}$ and an epistemic term 
$\mathbf g^\top \mathbf H^{-1}\mathbf g$, quantifying measurement noise 
and parameter uncertainty, respectively.

\section{Results}
\label{sec:results}
We evaluate performance on two tasks—\emph{jammer localization} (posterior inference of $\bftheta$) and \emph{field prediction} (RSS estimation at unobserved grid locations). Results include \emph{data generation}, \emph{quantitative analysis} across training sizes, and \emph{qualitative evaluation} of predicted fields and uncertainties.

\subsection{Data Generation and Model Details}
\label{sec:data}
\begin{figure}[t]
    \centering
    \includegraphics[width=0.9\linewidth]{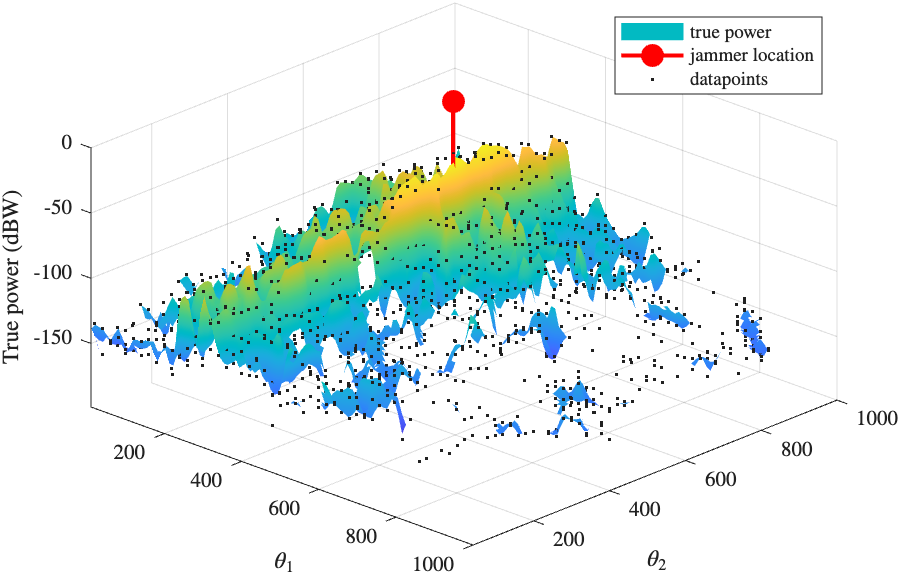}
    \caption{Simulated jammer RSS field for downtown Chicago. Urban canyon effects create higher power along aligned streets, while building shadows cause attenuation.}
    \label{fig:true_field}
\end{figure}

\textbf{Data Generation}. Synthetic data is commonly used in GNSS interference studies, as controlled real-world experiments with active jammers are both impractical and legally restricted. Following the methodology in~\cite{jaramillo-civill_jammer_2025,nardin_jamming_2023}, we adopt a Shooting and Bouncing Rays (SBR) ray-tracing approach to simulate the propagation field produced by a jammer in a $1\,\text{km}^2$ downtown Chicago area using building geometry from OpenStreetMap. The jammer transmits at $P_0=10$ dBW on the GPS L1 frequency (1575.42 MHz) with an isotropic antenna pattern. We focus exclusively on this urban scenario (Fig.~\ref{fig:true_field}) since urban propagation with multipath and shadowing presents the most challenging case~\cite{jaramillo-civill_jammer_2025,nardin_jamming_2023}. Despite the isotropic transmission, the resulting RSS field exhibits strong anisotropy due to urban canyon effects and building shadowing. The dataset is preprocessed to exclude points inside buildings, retaining only locations with potential GNSS coverage.

\noindent\textbf{Model Details.} 
The CNN expert (Fig.~\ref{fig:cnn_architecture}) takes $\bfX=[\mathbf{H}_{\text{bld}},\mathbf{P}_x,\mathbf{P}_y]\in\mathbb{R}^{3\times H\times W}$ and processes it through three convolutional layers with $3{\times}3$ kernels, batch normalization, ReLU activation, and dropout, producing feature maps with 16, 32, and 64 channels.  
A final $3{\times}3$ convolution and an average pooling layer project to a single output channel, yielding an $H{\times}W$ prediction of the RSS field.
The prior hyperparameters~\eqref{eq:prior_p0}--\eqref{eq:prior_omega} were set as $P_{\min}=5,\;P_{\max}=20$, $\gamma_{\min}=2,\;\gamma_{\max}=10$, $\sigma_c=10$, and $\tau=5$. To optimize the log-posterior in~\eqref{eq:log_posterior} we used the Adam optimizer~\cite{kingma_adam_2014}.

\subsection{Qualitative Results}
\begin{figure*}[htb]
    \centering
    \includegraphics[width=\linewidth]{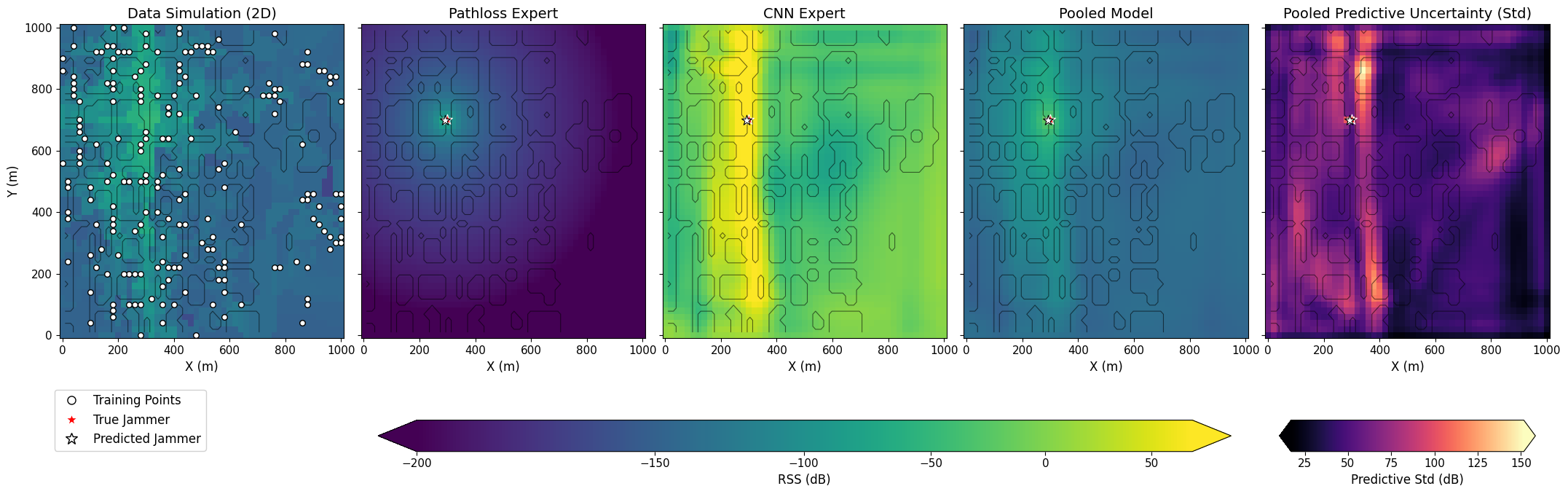}
    \vspace{-0.8cm}
    \caption{
        Five-panel visualization of jammer localization experiments. 
        From left to right: 
        (1)~Simulated RSS field with 200 training points overlaid; 
        (2)~Path-loss expert prediction $\mu^{\mathrm{PL}}(\mathbf p; \hat{\bfpsi}_{\mathrm{MAP}_{\bftheta,P_0,\gamma}})$; 
        (3)~CNN expert prediction $\mu^{\mathrm{CNN}}(\mathbf p; \bfX,\hat{\bfpsi}_{\mathrm{MAP}_{\bfomega}})$; 
        (4)~Pooled mean $\mu(\mathbf p;\mathbf{H}_{\text{bld}},\hat{\bfpsi}_{\mathrm{MAP}})$ (equals the mean of the predictive distribution under the Laplace approximation of the posterior); 
        and (5)~Pooled predictive uncertainty (standard deviation). 
        All panels include the building layout in the background and markers for the true jammer position (red star) and predicted position (white star). 
        Panels~(1)--(4) share a common RSS colorbar, while panel~(5) uses a separate uncertainty scale.
        }
    \label{fig:results}
\end{figure*}

Fig.~\ref{fig:results} provides a five-panel qualitative comparison of jammer localization experiments:  
(1)~the simulated RSS field using a total of 200 training points;  
(2)~the PL expert mean $\mu^{\mathrm{PL}}(\mathbf p; \hat{\bfpsi}_{\mathrm{MAP}_{\bftheta,P_0,\gamma}})$;  
(3)~the CNN expert mean $\mu^{\mathrm{CNN}}(\mathbf p; \bfX,\hat{\bfpsi}_{\mathrm{MAP}_{\bfomega}})$;  
(4)~the pooled mean $\mu(\mathbf p;\mathbf{H}_{\text{bld}},\hat{\bfpsi}_{\mathrm{MAP}})$ (under the Laplace approximation of the posterior, given by predictive distribution in~\eqref{eq:pred_dist}), and
(5)~the predictive standard deviation.  
Here, $\hat{\bfpsi}_{\mathrm{MAP}_{\bftheta,P_0,\gamma}}$ and $\hat{\bfpsi}_{\mathrm{MAP}_{\omega}}$ denote the subsets of the MAP estimate $\hat{\bfpsi}_{\mathrm{MAP}}$ corresponding to the path-loss and CNN parameters, respectively.  

The PL expert, guided by the centroid-based prior, helps steer the estimate toward the true jammer but yields a peaky field that poorly matches the propagation in complex realistic urban environments (as also noted in~\cite{jaramillo-civill_jammer_2025}). The CNN expert compensates for the PL’s rapid decay by elevating predictions along the street (urban canyon) of the jammer, where multipath increases the received power, while also capturing the sharp drops in other streets caused by blockage from surrounding buildings. The pooled mean inherits both effects, improving alignment with the true field. The predictive uncertainty is highest near the jammer and along narrow urban canyons, where small changes in position produce large RSS variations, and decreases in distant regions where the field is flatter and easier to extrapolate.

\subsection{Quantitative Evaluation}
\begin{figure}[t]
    \centering
    \includegraphics[width=1\columnwidth]{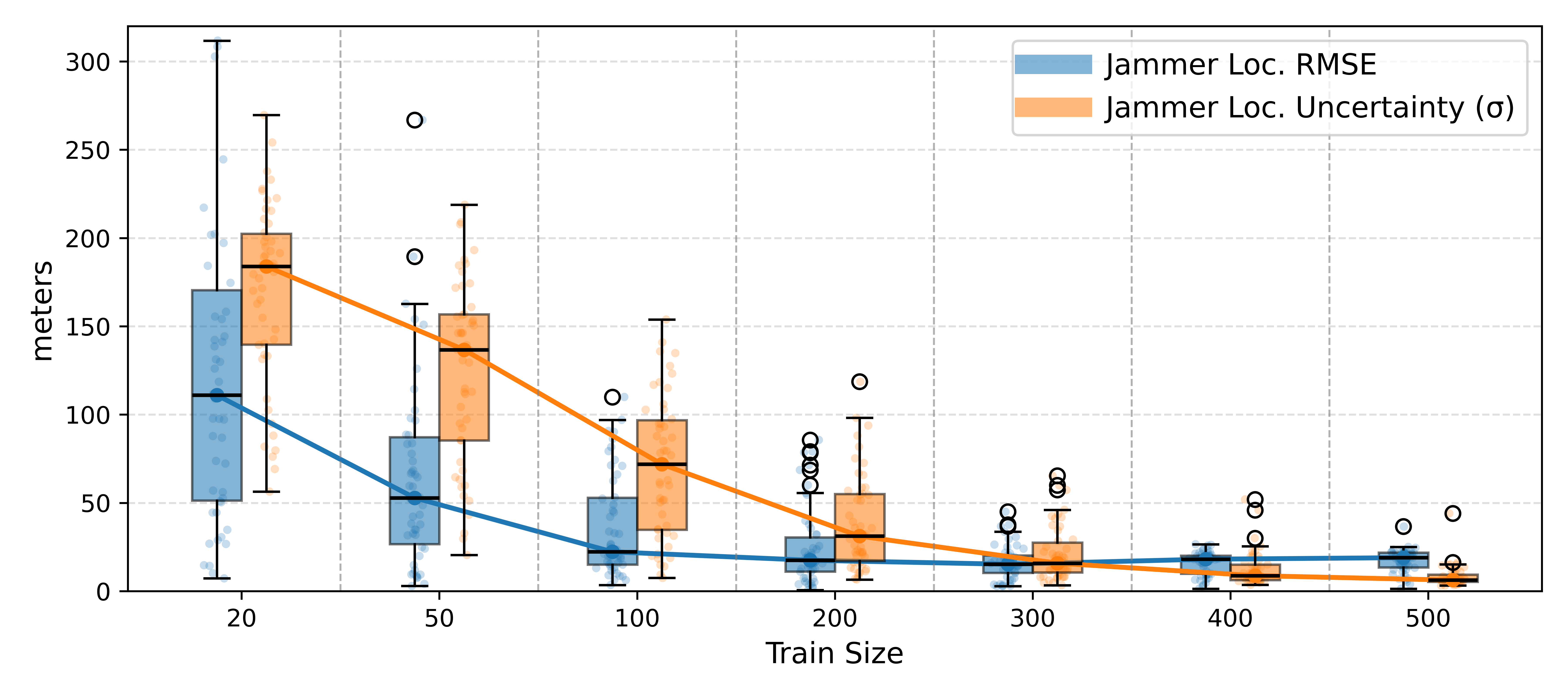}
    
    
    \includegraphics[width=1\columnwidth]{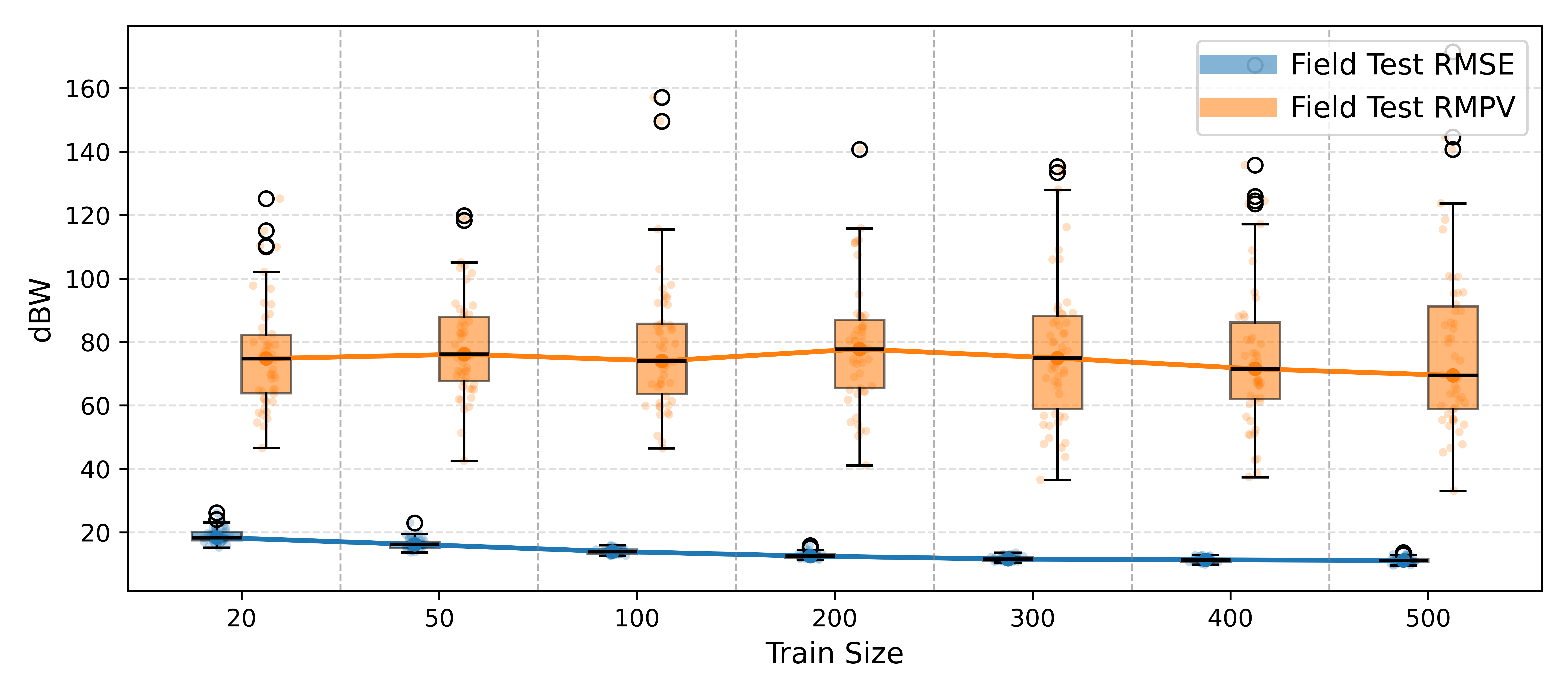}
    \vspace{-2.5em}
    \caption{
    Performance versus training set size (20–500 points), averaged over 50 Monte Carlo runs. 
    \emph{Top}: Jammer localization error (m) and posterior std. dev. (m).
    \emph{Bottom}: RSS field test RMSE (dBW) and field test RMPV (dBW).
    }
    \label{fig:train_size_analysis}
\end{figure}

Fig.~\ref{fig:train_size_analysis} shows performance versus training size (20--500 points), reporting boxplots across 50 Monte Carlo runs.

\noindent\textbf{Jammer localization.}
Median localization error decreases monotonically as the number of training samples increases, while the variance across runs also shrinks. This behavior arises because the data-dependent prior (temperature-weighted centroid) becomes increasingly informative with larger $N$. The posterior standard deviation obtained via the MAP-based Laplace approximation is initially conservative (underconfident) when data are scarce but becomes progressively better calibrated as the number of training points increases. For instance, at $N=200$, the empirical localization error is approximately $24$\,m, whereas the estimated uncertainty is about $38$\,m. We analyzed scatter plots of the localization errors, $\mathbf{e}_{\bftheta} = \hat{\bfpsi}_{\mathrm{MAP}_{\bftheta}} - \bftheta_{\text{TRUE}}$, and observed that the distribution deviates slightly from a Gaussian, which explains the mild underconfidence of the Laplace approximation that assumes local Gaussianity around the MAP estimate to make posterior inference tractable. As $N$ increases, the variance decreases, indicating that the posterior becomes sharper and more concentrated around the true jammer position.

\noindent\textbf{Field prediction.}
Test RMSE improves with $N$, indicating better generalization. We report the root mean predictive variance, $\mathrm{RMPV}=\sqrt{|\mathcal S|^{-1}\!\sum_{\mathbf p'\in\mathcal S}\mathrm{Var}[y'\mid \mathbf p',\mathcal D]}$ over test points $\mathcal S$, as an uncertainty indicator for field prediction. RMPV exceeds empirical RMSE (conservative), which is reasonable due to: (i)~sparse 2D supervision with building heights providing insufficient spatial context, (ii)~nonstationary spatial propagation in urban canyons, (iii)~Gauss--Newton approximation overestimating epistemic variance at high gradients~\cite{immer_disentangling_2020}. With building heights as the sole spatial context, distant predictions are poorly constrained, increasing predictive variance.

\section{Conclusion}
\label{sec:conclusion}
We presented a Bayesian mixture-of-experts model that fuses a path-loss and a CNN expert for jammer localization.  
Our formulation jointly infers the jammer position and reconstructs the propagation field while quantifying uncertainty in the propagation field.  
Experiments on urban ray-tracing data show that localization accuracy improves as training points increase and that the pooled hybrid model better captures the spatial structure of the RSS field.
We observe that uncertainty concentrates near the jammer and within urban canyons while diminishing in areas where the field is smoother and easier to extrapolate.  
Overall, the results demonstrate that the proposed hybrid framework effectively integrates physical modeling with data-driven spatial representations to achieve reliable and uncertainty-aware estimation of both the RSS field and jammer position.

\vfill \pagebreak
\clearpage

\bibliographystyle{IEEEbib}
\bibliography{ICASSP}
\end{document}